\documentclass[9pt,sigconf]{acmart} %

\usepackage{booktabs} 
\usepackage{CJK}
\usepackage{algorithmicx}
\usepackage[ruled,vlined,linesnumbered]{algorithm2e}
\usepackage{hhline,tabularx}
\usepackage{amsmath,mathtools}
\usepackage{amsfonts}
\usepackage{array}
\usepackage{mathrsfs}
\usepackage{hyperref}
\usepackage{graphicx}
\usepackage{subcaption}
\usepackage{enumitem,color}
\copyrightyear{2023}
\acmYear{2023}
\setcopyright{acmlicensed}\acmConference[SEC '23]{The Eighth ACM/IEEE Symposium on Edge Computing}{December 6--9, 2023}{Wilmington, DE, USA}
\acmBooktitle{The Eighth ACM/IEEE Symposium on Edge Computing (SEC '23), December 6--9, 2023, Wilmington, DE, USA}
\acmPrice{15.00}
\acmDOI{10.1145/3583740.3626612}
\acmISBN{979-8-4007-0123-8/23/12}

\begin{document}
\title[AdaMap]{\emph{AdaMap}: High-Scalable Real-Time Cooperative Perception at the Edge}


\author{Qiang Liu, Yongjie Xue, Yuru Zhang}
\affiliation{%
  \institution{University of Nebraska-Lincoln}
}
\email{qiang.liu@unl.edu}

\author{Dawei Chen, Kyungtae Han}
\affiliation{%
  \institution{Toyota InfoTech Labs}
}
\email{{dawei.chen1, kt.han}@toyota.com}


\renewcommand{\shortauthors}{Qiang Liu, et. al}

\begin{abstract}
Cooperative perception is the key approach to augment the perception of connected and automated vehicles (CAVs) toward safe autonomous driving.
However, it is challenging to achieve real-time perception sharing for hundreds of CAVs in large-scale deployment scenarios.
In this paper, we propose AdaMap, a new high-scalable real-time cooperative perception system, which achieves assured percentile end-to-end latency under time-varying network dynamics.
To achieve AdaMap, we design a tightly coupled data plane and control plane.
In the data plane, we design a new hybrid localization module to dynamically switch between object detection and tracking, and a novel point cloud representation module to adaptively compress and reconstruct the point cloud of detected objects.
In the control plane, we design a new graph-based object selection method to un-select excessive multi-viewed point clouds of objects, and a novel approximated gradient descent algorithm to optimize the representation of point clouds.
We implement AdaMap on an emulation platform, including realistic vehicle and server computation and a simulated 5G network, under a 150-CAV trace collected from the CARLA simulator.
The evaluation results show that, AdaMap reduces up to 49x average transmission data size at the cost of 0.37 reconstruction loss, as compared to state-of-the-art solutions, which verifies its high scalability, adaptability, and computation efficiency. 
\end{abstract}



\maketitle

\section{Introduction}
\label{sec:introduction}

Autonomous driving and advanced driving assistance system~\cite{yurtsever2020survey, wang2018networking,liu2020computing,ziebinski2017review} generally control the vehicle and react to diversified roadway environments based on the local perception obtained from the perspective of the ego vehicle.
To achieve robust perception, connected and automated vehicles (CAVs)~\cite{elliott2019recent,uhlemann2018time} are equipped with a variety of onboard sensors, such as cameras, LiDAR, and radar. 
However, these line-of-sight sensors can be contaminated under diverse roadway conditions~\cite{ahmad2020carmap}, e.g., vision occlusion, extreme weather, and sensor failures, which may lead to incomplete and inaccurate information in the local perception~\cite{liu2021edgesharing, liu2021livemap}.
In addition, the limited view angle of the ego vehicle might confuse and even mislead the local perception.
An example in~\cite{tesla-post} illustrates that a horse-drawn carriage is recognized as a truck or sedan intermittently by Tesla Autopilot. 
As a result, recent controversial accidents~\cite{stilgoe2020killed, vlasic2016self} of pilot commercial deployment raises widespread concerns about autonomous driving techniques, in terms of safety, reliability, and availability.

\begin{figure}[!t]
	\centering
	\includegraphics[width=3.34in]{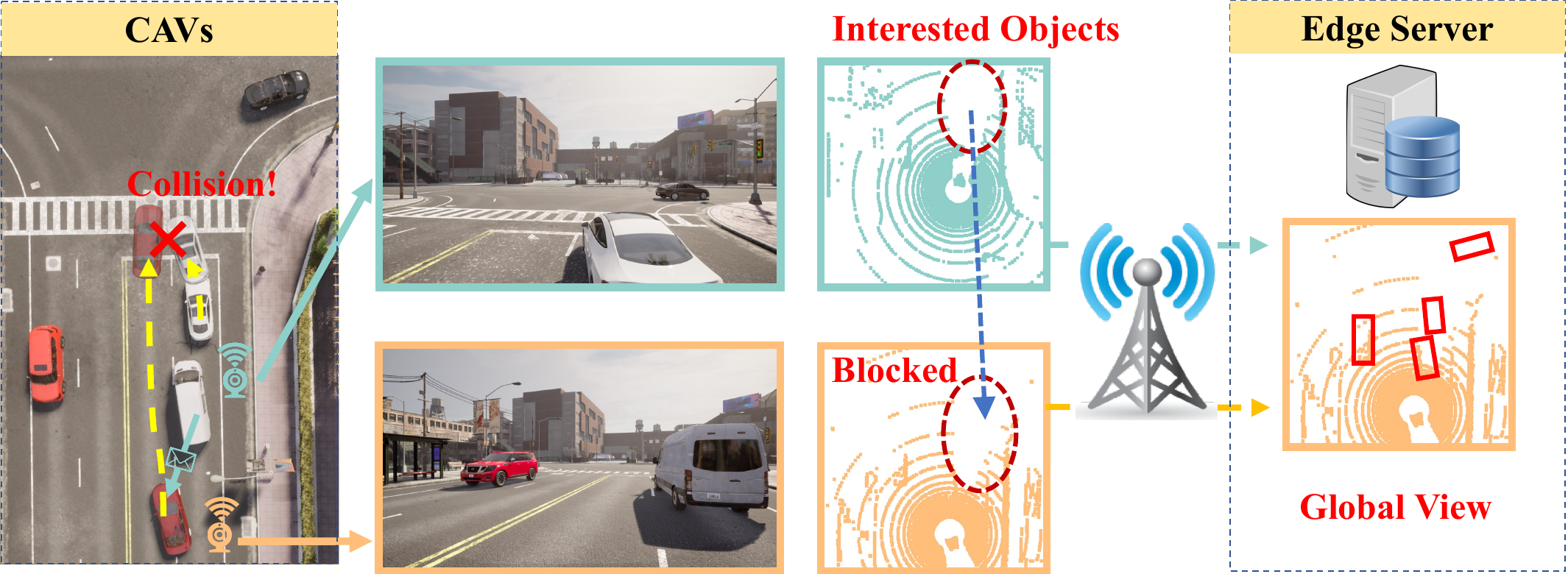}
	 \caption{\small An illustrative example of cooperative perception.}
	\label{fig:motivation}
\end{figure}

Cooperative perception~\cite{zhang2021emp, chen2019f, yu2021edge} has been extensively investigated to overcome the limitations of local perception in CAVs.
As illustrated in Fig.~\ref{fig:motivation}, cooperative perception allows the sharing of perception data (e.g., point clouds, images and locations) among vehicles (i.e., V2V~\cite{fukatsu2019millimeter}) and infrastructures (i.e., V2I~\cite{tsukada2020autoc2x}), such as lampposts and edge servers.
Thus, the ego vehicle can utilize the shared perception data to augment its local perception (e.g., reducing occluded areas and extending sensing ranges) for achieving high-safety driving~\cite{ahmad2020carmap, liu2021livemap}.
In particular, LiDAR point clouds are widely adopted in cooperative perception~\cite{chen2019f, qiu2022autocast}, which attributes to the unique property of privacy-preserving and weather-proof.
In cooperative perception, the key goal is to achieve \textit{real time sharing} for assuring the freshness of shared information~\cite{shi2022vips, hu2022where2comm, ahmad2020carmap}.
Given that commercial off-the-shelf LiDARs are with 10 Hz sampling rates, the end-to-end latency of perception sharing should be below 100 milliseconds, if not less.
The shared perception data with unsatisfied latency provides trivial information regarding roadway environments~\cite{liu2021livemap}, e.g., the location shift of 50mph vehicles can be more than 2 meters under 100ms delay.


\begin{figure}[!t]
	\centering
	\includegraphics[width=3in]{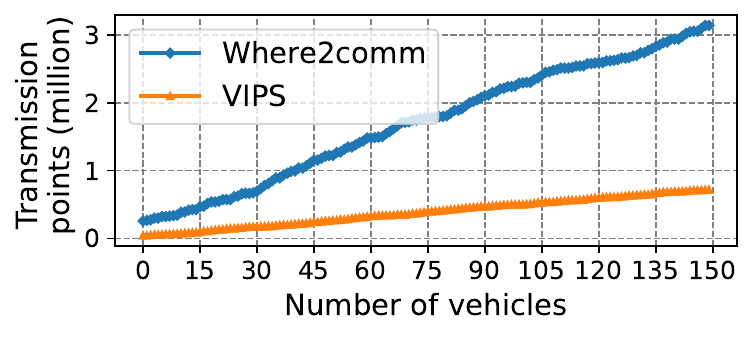}
	 \caption{\small Number of sharing point clouds under different vehicle traffic.}
	\label{fig:motivation_curve}
\end{figure}

It is, however, challenging to achieve real-time and large-scale cooperative perception, especially under time-varying network dynamics.
For example, given the 1+ million/s point clouds of commercial 32-thread LiDARs, the required wireless data rate to support 10 CAVs could reach up to 400 Mbps~\cite{shi2022vips}, which would lead to unsustainable network subscription fees for service providers.
Recent works~\cite{shi2022vips, hu2022where2comm, qiu2018avr} investigated transmitting only interested areas/objects (rather than the whole raw point cloud) to selective CAVs, and obtained very promising reductions in the over-the-air transmission data size.
However, existing works are focused on supporting a limited number of CAVs (e.g., less than ten), which still require considerable networking resources in large-scale deployment scenarios (e.g., 100+ CAVs). 
According to the forecast of Automotive Edge Computing Consortium (AECC), more than 50\% of cars on the road in the United States will have connected features by 2025~\cite{AECC}.
As a result, the cooperative perception system requires to support a massive number of CAVs, e.g., a 5G base station (1-mile radius coverage) may include hundreds of CAVs in metropolitan areas.
In Fig.~\ref{fig:motivation_curve}, we show the total number of points to be transmitted in existing works (i.e., VIPS~\cite{shi2022vips} and Where2comm~\cite{hu2022where2comm}) under different number of CAVs, based on our collected CARLA traces (see Sec.~\ref{sec:evaluation}).
To support the cooperative perception for 150 CAVs, Where2comm generates more than 3 millions points per frame, which needs to be transmitted over-the-air via wireless networks.
Moreover, existing works cannot adapt to time-varying network dynamics in automotive edge computing (e.g., radio channel and traffic variation), which thus fail to assure consistent end-to-end latency for real-time cooperative perception~\cite{shi2022vips, hu2022where2comm}.
Therefore, it is dispensable to investigate high-scalable cooperative perception with percentile latency assurance toward future large-scale deployment scenarios.


In this paper, we propose AdaMap, a new real-time cooperative perception system, that achieves high-scalable perception sharing with assured percentile latency under time-varying network dynamics.
The fundamental idea is two-fold. 
1) We adaptively represent the point cloud of objects to achieve higher compression ratios by trading off reconstruction losses.
This is based on the observation that, although reconstructed point clouds have perceivable reconstruction losses, they maintain sufficient structural details of point clouds regarding original objects.
2) We dynamically select objects detected by individual CAVs to reduce the total number of interested objects to be transmitted and processed.
This is based on the observation that, removing duplicated multi-viewed point clouds have only trivial degradation, regarding the aggregated fidelity of objects.
In AdaMap, we build an edge-based global map, which aggregates the uploaded objects from individual CAVs, monitors and tracks aggregated objects, and broadcasts periodical updates to update the local map in all CAVs.
Besides, by reusing the historical point clouds of objects, AdaMap could further reduce the transmission data size of all CAVs over time substantially.
In particular, we achieve AdaMap by designing a tightly coupled data plane and control plane.


The data plane of AdaMap is to process raw point clouds in individual CAVs into sharable object descriptors (e.g., location and point cloud of vehicles).
We design the data plane with two key modules.
First, we design a new hybrid object localization module to adaptively switch between object detection and tracking, which reduces 26x inference time at the cost of a minimal 0.02m localization error on average.
Second, we design a new flexible representation module to compress point clouds into low-dim latents based on variational autoencoder (VAE), which can achieve up to 64x compression ratio at the cost of 0.48 reconstruction loss on average.


The control plane of AdaMap is to adapt the data plane by optimizing the selection of detected objects and representation factor (RF) of selected objects in individual CAVs.
To address the unique optimization problem in AdaMap, we design the control plane to be compute-efficient with three key steps.
First, we decompose the problem into fully-distributed subproblems in individual CAVs, which mitigates communication overheads incurred in centralized optimization.
Second, we design a new graph-based object selection method to selectively remove duplicated multi-viewed point clouds, by evaluating the metric of space density.
Third, we design a new approximated gradient descent algorithm to optimize the RF of selected objects, which assures the percentile end-to-end latency under network dynamics.

We implement AdaMap on an emulation platform, which includes realistic vehicle and server computation and a simulated 5G network.
We finish more than 4000 lines of codes to implement AdaMap, which is ready-to-deploy in real-world vehicles and servers. 
We extensively evaluate AdaMap in terms of efficacy, scalability, and adaptability, under a CARLA dataset with 150 CAVs at 100 time frames. 
The results show that, AdaMap reduces up to 49x over-the-air transmission data size at the cost of only 0.37 reconstruction loss, as compared to state-of-the-art solutions.
Notably, AdaMap can achieve 89.4\% end-to-end latency within 100ms for 150 CAVs, given only 0.2MHz wireless bandwidth in 5G networks, which sheds light on large-scale cooperative perception deployment in the future.



\begin{figure}[!t]
  \centering
  \includegraphics[width=3.4in]{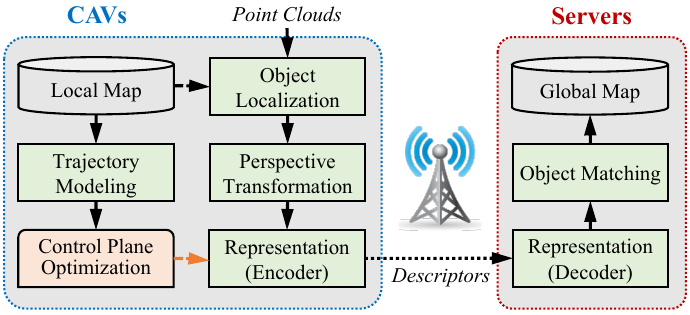}
  \caption{Overview of AdaMap.}
  \label{fig:overview}
\end{figure}

\section{System Overview}
The goal of AdaMap is to achieve high-scalable perception sharing among CAVs with assured percentile latency under time-varying network dynamics.
Besides, AdaMap needs to maintain high-fidelity objects regarding their aggregated point clouds, for further usage of autonomous driving pipelines, such as 3D scene understanding.
To achieve the goal, we design AdaMap with a tightly coupled data plane and control plane in a fully distributed manner, shown in Fig.~\ref{fig:overview}.
1) The object localization module localizes interested objects on the LiDAR point clouds, including 3D bounding boxes, orientations, and labels.
2) The perspective transformation module transforms detected objects from the local coordinate of CAVs into the global coordinate.
3) Then, the control plane applies to optimize the object selection and representation factor (RF) of selected objects in individual CAVs.
4) The representation (encoder) module encodes the point cloud of only selected objects into low-dim latents, according to the optimization results from the control plane. 
5) An object descriptor will be built for each selected object, where CAVs will send all object descriptors to their associated edge servers.
6) The representation (decoder) module at the edge server decodes received latents and reconstructs the point cloud of objects.
7) The object matching module matches and updates received objects within historical objects in the global map.
New updates of AdaMap will be periodically broadcasted to all CAVs.

\section{Data Plane Design}
\label{sec:data_plane}
In this section, we describe the data plane, including object localization, point cloud representation, perspective transformation, trajectory modeling, and object matching.

\subsection{Hybrid Object Localization}
\label{sec:object_localization}
Object localization module is to localize and extract interested objects (e.g., pedestrians and cars) from 3D point clouds perceived by the ego vehicle.
In AdaMap, we focus on only dynamic objects, e.g., moving cars and scooters, rather than stationary backgrounds (e.g., buildings and trees), which could be provided by high-definition (HD) maps.
Extracting only dynamic objects will substantially reduce the total data size of point clouds to the edge server~\cite{qiu2018avr}.
Given a point cloud, the output of object localization includes the 3D bounding boxes and the extracted point cloud of each detected object.


To localize objects in point clouds, there are generally two approaches, i.e., detection and tracking, can be used in AdaMap.
On the one hand, existing deep neural network (DNN) based object detection solutions~\cite{lang2019pointpillars,guo2020deep} are generally compute-intensive, which would greatly squeeze the end-to-end latency allowance for following data plane modules.
The state-of-the-art PointPillars~\cite{lang2019pointpillars} consumes 26.4 $\pm$ 4.7 ms to process point clouds in our CARLA dataset under NVIDIA RTX 3090 and Intel i7 CPU, which already occupies more than a quarter of the 100ms end-to-end latency allowance in AdaMap.
On the other hand, existing object tracking solutions can achieve real time inference, while their tracking accuracy varies under different roadway situations, such as acceleration and braking (see Table~\ref{tab:object_localization} and Fig.~\ref{fig:localization_error}). 
For example, Kalman filter~\cite{welch1995introduction} has been widely used for tracking in various application domains, where its inference time can be as low as sub-milliseconds.



\begin{figure}[!t]
  \centering
  \includegraphics[width=3.4in]{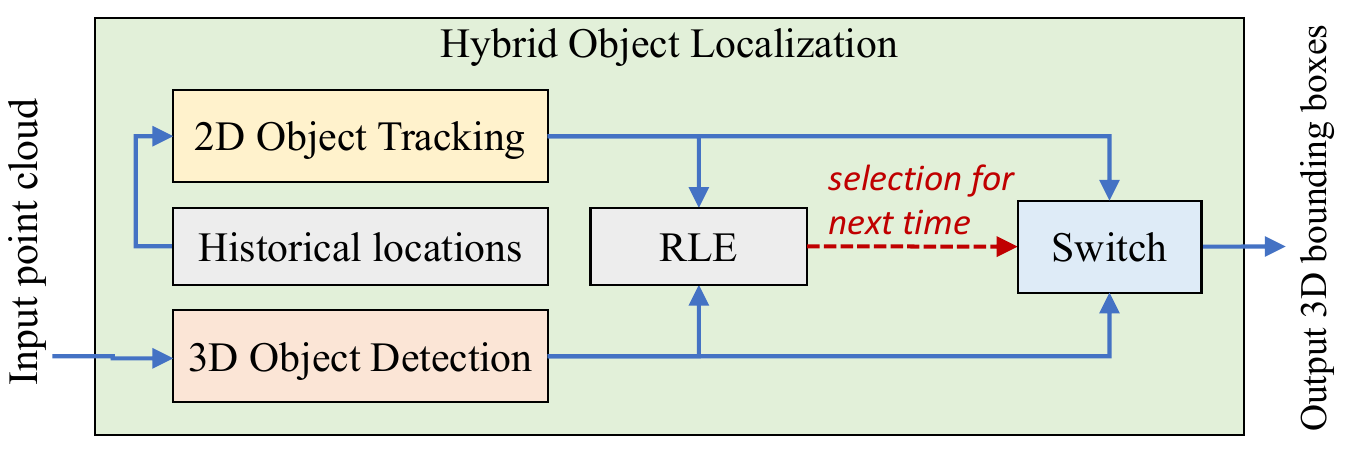}
  \caption{The hybrid object localization.}
  \label{fig:object_localization}
\end{figure}

To this end, we propose a new hybrid object localization module, as shown in Fig.~\ref{fig:object_localization}, to reduce the inference time at the cost of a minimal degradation of localization error.
The basic idea is to adaptively switch between object detection and tracking, where the inference time can be considerably reduced by invoking object tracking at runtime.
This is based on the observation that, object tracking can achieve considerable accurate localization in most cases in AdaMap.
On the one hand, the location of objects are aggregated in the object matching module (see Sec.~\ref{sec:object_matching}), which helps to correct the localization error in individual CAVs. 
On the other hand, the location of objects are updated frequently (e.g., maximum 100ms latency), where fine-grained historical locations help to improve the accuracy of trajectory modeling (see Sec.~\ref{sec:trajectory_modeling}).
However, it is challenging to determine when to switch between object tracking and detection.
Because we do not have the ground-truth of object locations in real-world deployment, and thus it is difficult to evaluate the localization accuracy of object tracking.

\begin{table}[!t]
  \centering
  \captionof{table}{Comparison among object localization methods}
  \begin{tabularx}{\linewidth}{*{3}{>{\centering\arraybackslash}X}}
    \toprule
       Method & localization error (m) &  inference time (ms) \\ \midrule
    Tracking  &   0.12$\pm$ 0.16 &    0.73 $\pm$ 0.71\\
    Detection  &   0.07 $\pm$ 0.04 &   26.41 $\pm$ 4.73 \\
    Ours &  0.09 $\pm$ 0.05 &    1.02 $\pm$ 0.98\\
    \bottomrule
  \end{tabularx}
  \label{tab:object_localization}
\end{table}

To solve this issue, we define a new metric, named relative localization error (RLE), as the Euclidean distance between the location derived by the object tracking and detection.
In other words, we consider the location generated by object detection as the ground-truth at runtime.
In each time slot, we calculate the relative localization error (RLE) based on the derived locations of detection and tracking.
The input of object detection and tracking are the whole raw point cloud and historical locations of previous detected objects, respectively.
Then, we determine which localization method will be used, for the next time slot.
For example, we select to use object tracking if the RLE is below a given threshold (empirically set as 0.5m in AdaMap).
In the next time slot, the object tracking will be invoked to derive the output of 3D bounding boxes, which will hide the inference time of object detection.
Otherwise, the object detection will be invoked at runtime, where its inference time will be included in the data plane of AdaMap.
Note that, object detection and tracking will be performed in parallel at any time slots.
For example, if we decide to use object tracking in the next time slot, the object detection will still be needed to derive the ground-truth location and calculate the RLE.


In Table.~\ref{tab:object_localization}, we show the localization error and inference time under different object localization methods.
As compared to object detection (i.e., PointPillars), our method reduces 96.1\% inference time at the cost of only additional 0.02m localization error.
As compared to the object tracking (i.e., Kalman Filter), our method reduces 25\% localization error, while the inference time is only increased 0.29ms on average.
These results show that, our proposed hybrid localization methods can substantially accelerate the object localization at the cost of negligible degradation on localization losses.

\subsection{Flexible Point Cloud Representation}

The representation module is to represent point clouds by using low-dim vectors, where the vectors should be able to accurately recover the original point clouds afterward.
In AdaMap, the information of objects (e.g., locations and point clouds) will be uploaded to edge servers and shared with all CAVs, where point clouds dominate the overall data size of object descriptors (see Sec.~\ref{sec:object_descriptor}).
Although the data size of point clouds of individual objects are relatively small (e.g., a few Kbits for 1000 points), the aggregated data size of thousands of objects (if not more) will require considerable networking resources.
General-purpose compression solutions (e.g., Zstandard~\cite{collet2018zstandard}) can be used to compress and decompress point clouds at runtime.
However, the achieved compression ratios are typically limited for detected objects in AdaMap (e.g., 1.91 $\pm$ 0.06 for Zstandard), which mainly attributes to the sparse point clouds (e.g., 992 $\pm$ 1206 points in our CARLA dataset) of detected objects.

\begin{figure}[!t]
	\centering
	\includegraphics[width=3.4in]{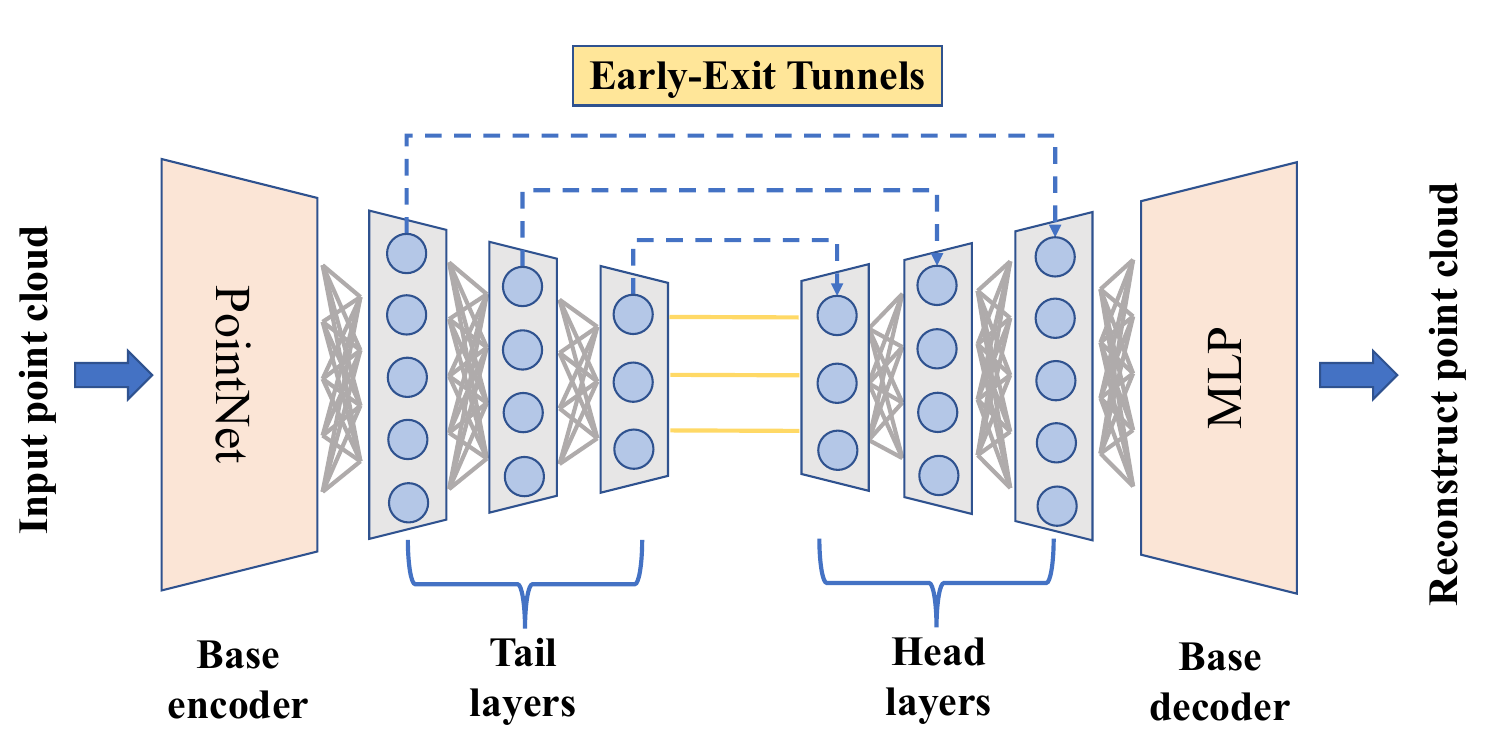}
	\vspace{-0.05in} 
    \caption{\small The representation module with early-exit tunnels.}
	\label{fig:vaestructure}
\end{figure}

To this end, we propose a new flexible point cloud representation module based on variational autoencoder (VAE)~\cite{kingma2019introduction}.
As an unsupervised learning technique, an autoencoder generally includes an encoder that compresses the input data (e.g., point cloud) into low-dim latents with a fixed length, and a decoder that reconstructs the original data from the latents.
Different from general-purpose compression solutions, autoencoders show great potentials to achieve high compression ratios under low reconstruction losses~\cite{liu2021livemap}, for seen point clouds in the training dataset.
However, the reconstruction loss might be very high for unseen point clouds (e.g., new car models), as they are not in the training dataset.

In AdaMap, we design the representation module based on VAE for the following reasons.
First, we can obtain the structure model of almost all vehicles on the market and collect their point clouds in advance to be a complete training dataset.
Because the majority of vehicles on the road are with specific models defined by their manufacturers.
With the complete training dataset, the VAE will be trained to obtain minimal reconstruction losses, which assures the accurate reconstruction of point clouds in edge servers\footnote{With a complete training dataset, partial-view point clouds can be represented and even completed with a high quality and accuracy~\cite{liu2020morphing,yu2021pointr}.}.
Second, we adopt VAE rather than conventional autoencoder, where the generated latents of VAE serve as not only the low-dim representation but also the feature of point clouds in the latent space~\cite{kingma2019introduction, liu2021livemap}.
By introducing a regularization term in the loss function, VAE regularizes the latent space, where similar input point clouds would generate similar latents.
These unique features would be useful for further improving AdaMap, such as increasing matching accuracy via feature-aware object matching and enhancing the generalization of representation via adversarial training (see Sec.~\ref{sec:discussion}).
However, we found that using VAE with fixed-length latents (i.e., fixed compression ratios) cannot adapt to time-varying network dynamics in AdaMap.
In particular, CAVs usually experience varying radio channel qualities under high-speed mobility, where their non-stationary volatile wireless data rates would lead to fluctuating radio transmission latency.

To address this issue, we design an flexible representation module with multiple early-exit tunnels between the encoder and decoder.
As shown in Fig.~\ref{fig:vaestructure}, the representation module consists of a base encoder followed by multiple tail layers, and multiple head layers followed by a base decoder.
We denote an \textit{early-exit tunnel} as the pair of a tail and a head layer with the corresponding dimension of latent space.
For example, we may create 256x128x64 tail layers and 64x128x256 head layers, where three early-exit tunnels will be enabled, including the outermost (256 dim), middle (128 dim), and innermost (64 dim).
The innermost early-exit tunnel generates latents with the lowest dimension (i.e., the highest compression ratio), and obtains the highest reconstruction loss.
In contrast, the outermost early-exit tunnel obtains the lowest compression ratio and the reconstruction loss in the meantime.
With this unique design, our proposed flexible representation module allows dynamic configuration of representation factor (RF) at runtime, without the need for retraining.
Here, we define \textit{representation factor (RF)} as the ratio between the dimension of input point clouds and the generated latents under individual early-exit tunnels in the representation module. 

\begin{table}[!t]
\renewcommand{\arraystretch}{1.5} 
\setlength{\tabcolsep}{4.3pt} 
\caption{Reconstruction loss under different RFs}
\label{tab:loss_under_rf}
\begin{tabular}[b]{c c cccc}\hline
  RF   & 4 & 8 & 16 & 32 & 64\\ \hline
  \textbf{Loss} & 0.26 $\pm$ 0.1 & 0.32 $\pm$ 0.1 & 0.38 $\pm$ 0.3 & 0.46 $\pm$ 0.3 & 0.48 $\pm$ 0.4\\ \hline
\end{tabular}
\label{tb:network_performance_comparison}
\end{table}


\begin{figure}[!t] 
  \begin{minipage}[t]{0.235\textwidth}
    \centering
    \includegraphics[width=\textwidth]{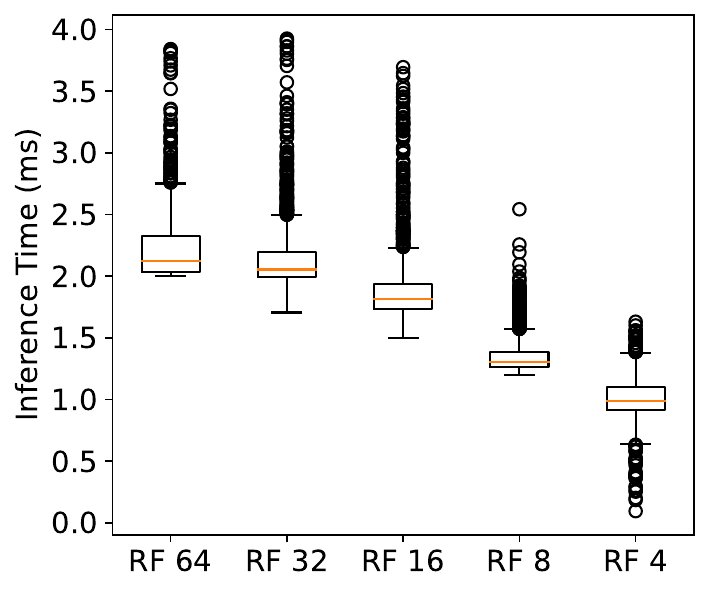}
    \captionof{figure}{\small Inference time of representation module.}
    \label{fig:RF_vs_latency}
  \end{minipage}
  \hfill
  \begin{minipage}[t]{0.235\textwidth}
    \centering
    \includegraphics[width=\textwidth]{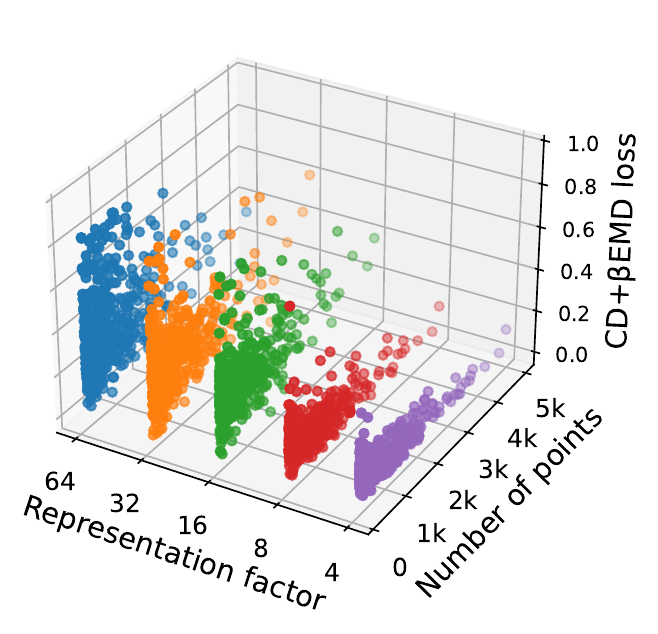}
    \captionof{figure}{\small Reconstruction loss of representation module.}
    \label{fig:RF_context}
  \end{minipage}
\end{figure}

Next, we design the process for training the proposed representation module.
First, we create the VAE to include only the base encoder, the outermost early-exit tunnel (e.g., 256-dim tail and head layer), and the base decoder.
Second, we train the VAE extensively and freeze all of its parameters.
Third, we add an additional early-exit tunnel, and train the VAE by updating only the parameters of the newly added tunnel, such as the 128-dim tail and head layer. 
We repeat the second and third steps to add more early-exit tunnels toward the final representation module.
Note that, the representation module cannot be trained as a whole (e.g., in one training process), because it will break the early-exit mechanism, where the generated latents will not be able to recover the original point clouds in the decoder, except the innermost tunnel.

In addition, we design the loss function of the representation module to include the regularization loss and the reconstruction loss. 
First, the regularization loss is defined as $D_{KL} \left[ q(z|x) | p(z)\right]$, where $x$ are the input point clouds and $z$ are the sampled latents.
Here, $D_{KL}$ is the KL-divergence to evaluate the difference between the probability distribution of latents and a prior distribution (Gaussian $p(z) \sim \mathcal{N}(0,1)$).
Second, the reconstruction loss is defined as the weighted sum of Chamfer Distance (CD)~\cite{wu2021balanced} and EarthMover’s Distance (EMD)~\cite{wen2021pmp}, which are the widely used metrics to evaluate the difference between point clouds.
Specifically, the metric of CD and EMD are mainly focused on the local and global scale, respectively.
This is based on our observation that, using merely CD or EMD will result in significant loss on global and local details in reconstructed point clouds, respectively.
Hence, the reconstruction loss is expressed as 
\begin{equation}
\label{eq:reconstruction_loss}
    loss = Loss_{CD} + \beta \cdot Loss_{EMD},
\end{equation}
where $\beta$ is the non-negative weight factor.

Table~\ref{tab:loss_under_rf} shows the reconstruction loss achieved by the representation module under different RFs, where all input point clouds are re-sampled into 1024 points.
We can see the tradeoff between compression ratios and reconstruction losses, where a higher RF generally leads to a higher reconstruction loss.
As compared to the average 0.26 reconstruction loss under RF 4, an additional 0.22 loss can trade for 16x more compression ratio under RF 64. 
Fig.~\ref{fig:RF_vs_latency} shows the inference time of the representation module under different RFs.
As more tail and head layers need to be computed for larger RFs, their inference times increase moderately.
Note that, although one-time inference of the representation module is minimal, the aggregated computation time for all objects in individual CAVs is non-negligible.

\subsection{Perspective Transformation}


The perspective transformation module is to transform the objects in the local coordinate of individual CAVs to the global coordinate system in the edge server.
We denote the 3D location of an object $P_l = [X,Y,Z,1]$ in the local coordinate.
Hence, the object's location in the global coordinate $P_g$ can be calculated as $P_g = P_l \; \times \; T_{l2g}$, where $T_{l2g}$ is the 4x4 transformation matrix defined as
\begin{equation}
T_{l2g} = \left[
    \begin{array}{cccc}
    c_{p} c_{y} & -c_{r} s_{y}+c_{y} s_{p} s_{r} & c_{r} c_{y} s_{p}+s_{r} s_{y} &X\\
    c_{p} s_{y} & c_{r} c_{y}+s_{p} s_{r} s_{y} & c_{r} s_{p} s_{y}-c_{y} s_{r} &Y\\
    -s_{p} & c_{p} s_{r} & c_{p} c_{r}&Z\\
    0&0&0&1
    \end{array}
\right],
\end{equation}
where $c$ and $s$ denote sine and cosine operations, and $p,r,y$ are the pitch, roll, and yaw, respectively.
At runtime, the transformation matrix could be obtained by using high-precision GPS and simultaneous localization and mapping (SLAM) techniques~\cite{mur2015orb}.

\subsection{Trajectory Modeling}
\label{sec:trajectory_modeling}
The trajectory modeling module is to model and predict the future path of dynamic objects (e.g., vehicles, scooters and pedestrians), based on their historical locations.
In AdaMap, the trajectory modeling will be utilized to track all identified objects in both the object localization module (see Sec.~\ref{sec:object_localization}) and object matching module (see Sec.~\ref{sec:object_matching}).
Although there are extensive solutions developed for trajectory modeling, e.g., deep learning-based~\cite{altche2017lstm, alahi2016social}, they generally have high computation complexity and thus lead to extra computation latency at runtime.
In AdaMap, we achieve trajectory modeling by using Kalman filter, which has proven robustness and computational efficiency under various application scenarios, such as robotics, UAVs, and aerospace.

The Kalman filter comprises two primary stages, namely the prediction stage and the correction stage. 
In the prediction stage, it uses the previous state estimation as the input to generate a forecast of the current state. 
In the correction stage, the predicted state obtained in the prediction stage is adjusted based on the current observed state. 
Without loss of generality, we define the state as $[X,Y,d_{X},d_{Y}]$, where $X$ and $Y$ are the current location of the CAV, and $Z$ is omitted.
The $d_{X}$ and $d_{Y}$ are the velocity of the CAV, which are initialized as zeros. 

\subsection{Predictive Object Matching}
\label{sec:object_matching}
The object matching module is to match multi-viewed objects received from individual CAVs, combine and update them into individual objects in the global map.
In AdaMap, objects might be detected intermittently under complex roadway environments, such as miss-detection in object detection and vision blocking.
Besides, all modules in the data plane might introduce noises and errors in real-world deployment~\cite{shi2022vips}, such as inaccurate detection and tracking, and transformation matrix error.
Hence, it is non-trivial to accurately match multi-viewed dynamic objects under large-scale deployment scenarios.

To this end, we develop a predictive-based matching method by evaluating the predictive location, rather than last-observed location, of objects at runtime.
Specifically, we predict the location of all objects in the global map by using the trajectory modeling module at each time frame.
Denote $g$ as the observed geo-location of an object in the world coordinate, we match it with the object $i =\arg\min_{k \in \mathcal{K}}\left\|g-\hat{g}_{k}\right\|^{2}$, where $\hat{g}_k$ is the predicted location of the $k$th object in the global map.

\subsection{Object Descriptor}
\label{sec:object_descriptor}
In AdaMap, we build the structure of object descriptor to include all the sharable information regarding an individual object to be transmitted.
To describe an object, an object descriptor incorporates the following attributes, i.e., location, orientation, 3D bounding box, label, confidence, speed, trajectory model, latents, and certain meta-data, such as the number of raw point cloud and time stamp.
The global map is mainly composed of object descriptors for all objects.
For the periodical updates, only changed attributes will be included in the object descriptor and broadcasted to all CAVs.


\section{Control Plane Design}
In this section, we describe the control plane, including system model, problem formulation, and solution design.


\subsection{System Model}
\label{sec:system_model}
We consider a generic mobile network, including multiple connected and automated vehicles (CAVs), wireless base stations (BSs), and edge computing servers.
As described in the data plane in Sec.~\ref{sec:data_plane}, CAVs periodically (e.g., 10Hz) upload the descriptor of detected objects to their associated edge servers in a time-slotted manner.
Edge servers will be responsible for further post-processing of object descriptors (e.g., the decoder representation module), in a first-come-first-serve manner.
Note that, we design the control plane of AdaMap from the perspective of service providers, which means that network transmission generally becomes a blackbox\footnote{Although network slicing techniques~\cite{liu2020edgeslice} may create a network slice with isolated resources to support AdaMap, it is impractical to assign exact radio resources to individual CAV transmissions in the time scale of subseconds.}.
Without loss of generality, we consider the association between CAVs and BSs and edge servers are pre-determined, and focus on adapting the data plane towards diversified networking and computation dynamics in AdaMap.

For the $i$th CAV, we denote $\mathcal{O}_i= \{o_1^i, o_2^i,..., o_{K_i}^i\}$ as the set of detected objects, where $K_i$ is the number of total detected objects of the $i$th CAV.
Denote $\mathcal{A}_i = \{a_1^i, a_2^i,...,a_K^i\}$ as the set of selection indicator.
Note that, $a_k^i=1$ means the $k$th detected object in the $i$th CAV is selected to be transmitted to edge servers, otherwise the object will not be transmitted.
Furthermore, we denote the set of representation factor (RF) of the $i$th CAV as $\mathcal{R}_i= \{r_1^i, r_2^i,..., r_{K_i}^i\}$, where $r_k^i$ is the RF for the $k$th object.
The action space of RFs are constrained by the minimum $r_{\min}$ and maximum $r_{\max}$, defined by the representation module in the data plane.
For example, a point cloud will go through the encoder and all the followed tail layers, if its RF is the maximum $r_{\max}$ to generate the most condensed latents.

\textbf{Fidelity Model.}
In AdaMap, we define the fidelity of an object as the difference between its input and reconstructed point cloud.
Without loss of generality, we measure the difference by mostly adopting the loss function of the representation module in the data plane. 
Specifically, given the discrete early-exit tunnels in the representation module, the fidelity is defined as
\begin{equation}
\label{eq:fidelity}
    f(r | s) = - (Loss_{CD} + \beta \cdot Loss_{EMD}),
\end{equation}
where $r$ is the RF of the object and $s$ is the auxiliary state.
The $Loss_{CD}$ and $Loss_{EMD}$ are the loss of CD~\cite{wu2021balanced} and EMD~\cite{wen2021pmp}, respectively.
The rationale of introducing the auxiliary state is based on the following observation.
In Fig.~\ref{fig:RF_context}, we observe that the reconstruction loss of individual objects relate to not only their RFs but also the total number of their raw point clouds.
Given a fixed RF, the resulted fidelity are largely distributed, but can be generally determined by the total number of raw point cloud.
In other words, the averaged fidelity regardless the auxiliary state $s$ cannot provide the effective guidance when optimizing individual objects in CAVs.

Therefore, the overall fidelity of AdaMap is defined as 
\begin{equation}
    F = \sum\limits_{i=1}^{N}{\sum\limits_{k=1}^{K_i} \alpha_k^i \cdot f(r_k^i | s)}.
\end{equation}

\textbf{Latency Model.}
The end-to-end latency is defined as the elapsed time of completing the whole data plane for individual CAVs, including vehicle-side processing, radio transmission, and server-side processing.
Under the complex and unpredictable network and computing dynamics, the data plane is with probabilistic transmission and computation time.
In particular, we conduct an experiment to measure the computation time of the representation module under more than 10K objects.
As shown in Fig.~\ref{fig:RF_vs_latency}, the computation time of the representation module under all RFs is 1.72 $\pm$ 0.53ms.
Based on the above observation, it is difficult to derive a deterministic formulation for end-to-end latency in AdaMap.

To accurately represent the uncertainty in AdaMap, we build the latency model by incorporating multiple probabilistic parts as
\begin{equation}
    L_i = \sum\limits_{k=1}^{K} {\alpha_k^i \cdot \left( \frac{c_v(r_k^i)}{R_v^i} + \frac{d(r_k^i)}{R_w^i} + \frac{c_e(r_k^i)}{R_e^i} + B_i\right)},
\end{equation}
which includes four parts, i.e., vehicle-side processing, radio transmission, server-side post-processing, and aggregated latency $B_i$.
Here, we focus on the optimization of RF and object selection in the control plane, so the incurred latency of all other modules are aggregated, such as detection and tracking in the CAV, and object matching in the edge server. 
The first vehicle-side processing latency is defined as the ratio between the computation complexity of the encoder representation module $c_v(r_k^i)$ and the computation capacity of the CAV $R_v^i$.
The second radio transmission latency is defined as the ratio between the size of generated latents $d(r_k^i)$ and the experienced wireless data rate $R_w^i$.
Note that, we design an approximated model for this radio transmission latency. 
This is because, we are optimizing AdaMap from the perspective of service providers who cannot control network transmission (e.g., radio resource allocation~\cite{liu2021onslicing,liu2021constraint}) and server computation.
The third server-side processing latency is defined as the ratio between the computation complexity of the decoder representation module $c_e(r_k^i)$ and the experienced computation capacity of the CAV $R_e^i$.
Note that, 1) non-optimization variables (i.e., $R_v^i, R_e^i$) and static functions (i.e., $c_v(r_k^i), d(r_k^i), c_e(r_k^i)$) can be profiled under frozen implementation of the representation module; 2) the end-to-end latency is probabilistic, even under the given RF $r_k^i$. 

\subsection{Problem Formulation}
In the control plane design, the objective is to maximize the expectation of overall fidelity while meeting the percentile requirement of end-to-end latency.
To achieve the goal, we focus on optimizing the selection and RF of detected objects in all CAVs.
Therefore, we formulate the control plane problem as 
\begin{align}
    \mathbb{P}_0: \;\;\; \max \limits_{\mathcal{A}_i, \mathcal{R}_i, \forall i} & \;\;\;\;\;  \mathbb{E}[F]  \\ 
     s.t. & \;\;\;\;\; Prob(L_i \le H) \ge p, \forall i \label{eq:const_latency} \\
      & \;\;\;\;\; r_k^i \in [r_{\min}, .., r_{\max}], \forall i, k  \label{eq:const_action_r} \\
      & \;\;\;\;\; \alpha_k^i \in [0, 1], \forall i, k, \label{eq:const_action_a} 
\end{align}
where $\mathbb{E} [\cdot]$ is the operator of expectation.
Besides, $H$ is the given latency requirement (e.g., 100ms) and $p$ is the percentage threshold, such as 90\%.
The constraint in Eq.~\ref{eq:const_latency} assures that there are at least $p$ percentage CAVs have less than $H$ end-to-end latency.

\textbf{Challenges.}
The challenges of solving the above problem is multi-fold.
First, the problem has to be solved in a fully distributed approach.
In existing centralized approaches, the optimization is completed in a central server, where non-trivial communication delay and bandwidth overheads are incurred, especially when supporting large-scale geo-distributed users. 
In the distributed scenario, global information (e.g., locations and wireless data rate) is generally absent when conducting the optimization in individual CAVs.
Second, the problem involves multiple probabilistic parts in both the objective function and constraints, which falls into the field of stochastic programming.
Due to the uncertainty in the problem, existing optimization methods (e.g., gradient descents~\cite{boyd2004convex}) cannot be directly applied. 
Third, the distributed optimization needs to be compute-efficient for runtime decisions.
The transmission and computation in the data plane can only be proceeded until the optimization decision is derived. 
Given the stringent end-to-end latency requirement, the optimization time itself has to be minimal, such as less than 10ms.

\subsection{Algorithm Design}

In this section, we design a new control algorithm to effectively solve the problem $\mathbb{P}_0$ as follows.
First, we decompose the problem into multiple fully-distributed subproblems, where each subproblem is focused on optimizing the selection and RF of objects in individual CAVs.
The rationale of problem decomposition lies in the observation that, only two non-optimization variables (i.e., wireless data rate $R_w^i$ and server computation capacity $R_e^i$) couple the problem $\mathbb{P}_0$.
These subproblems will be addressed independently in corresponding CAVs at runtime.
Second, we optimize the object selection to minimize the total number of selected objects without compromising the overall fidelity.
The rationale behind is that, objects are generally viewed by multiple CAVs, where removing excessive point clouds will not decrease its aggregated fidelity perceivably.
Third, we optimize the RF for only selected objects to meet the percentile latency constraint based on Lagrangian primal-dual method.
To deal with the uncertainty in the latency and fidelity models, we design an approximated gradient descent method in solving the inner primal subproblem, inspired by the sample average approximation (SAA)~\cite{kim2015guide}.

\subsubsection{Problem Decomposition}
\label{sec:problem_decomposition}
We observe that the problem $\mathbb{P}_0$ is mainly coupled by the experienced wireless data rate $R_w^i, \forall i$ and server computation capacity $R_e^i, \forall i$, which are not optimization variables.
Given the blackbox property of radio transmission from the perspective of service providers, the future wireless data rate needs to be predicted no matter in the server (centralized) or vehicle (distributed) sides.
Besides, the first-come-first-serve edge servers can hardly be accurately managed and modeled in terms of the experienced computation capacity of individual CAVs, which generally leads to uncontrollable blackbox. 
Hence, we decompose the problem $\mathbb{P}_0$ into fully distributed subproblems in individual CAVs.
Without loss of generality, the subproblem in the $i$th CAV is expressed as
\begin{align}
    \mathbb{P}_1: \;\;\; \max \limits_{\mathcal{A}_i, \mathcal{R}_i} & \;\;\;\;\;  \mathbb{E}[F_i]  \label{eq:objective_function_individual}\\ 
     s.t. & \;\;\;\;\; Prob(L_i \le H) \ge p, \;\; \label{eq:const_latency_individual} \\
     & \;\;\;\;\; (\ref{eq:const_action_r}), \;\; (\ref{eq:const_action_a}), \nonumber
\end{align}
where the experienced wireless data rate $R_w^i$ and server computation capacity $R_e^i$ will be predicted at runtime in CAVs.


\subsubsection{Graph-Based Object Selection}
\label{sec:object_selection}
Here, we focus on optimizing the selection of detected objects in individual CAVs.
The goal is to un-select excessive multi-viewed point clouds of CAVs as many as possible, which corresponds to reducing the total number of object descriptors to be transmitted over-the-air and processed in edge servers.
The challenge of object selection lies in the complex aggregated fidelity, i.e., the fidelity after post-processing (e.g., registration, fusion) of multi-viewed point clouds of CAVs.
In other words, it is difficult to evaluate the impact of aggregated fidelity if any multi-viewed point clouds are removed.


To tackle this issue, we design a new graph-based selection method to selectively remove multi-viewed point clouds of objects.
In particular, we introduce a metric (named \textit{space density}), which is defined as the density of point clouds in a given 3D space. 
The proposed selection method iterates to remove objects, as long as the space density of objects are larger than the given threshold.
The rationale behind is that, excessive point clouds on the limited surface contributes negligible improvement on the fidelity of objects.
Because non-penetration LiDAR sensors can only scan the surface of objects, excessive point clouds will densely overlay on the surface without providing new structural information.
In AdaMap, the multi-viewed point clouds of the identical object will be registered and stitched together, where overlayed points will be mostly combined at the end.

The proposed object selection method is achieved as follows.
First, we partition the 3D bounding box of detected objects into multiple 3D sub-spaces, such as 1$m^3$ cubes.
Because LiDARs in CAVs can only perceive partial of an object (i.e., less than 180 degrees).
For example, we can consider that two multi-viewed point clouds of an object are both on the same side.
If we use the 3D bounding box of the object as a whole space, the object selection method would un-select all other multi-viewed point clouds, which will lead to imbalance space density.
In AdaMap, we partition the 3D bounding box of objects into 4 sub-spaces.
Second, for each 3D sub-space, we need to obtain the number of point cloud of objects for comparing with the given threshold.
However, we do not have the exact numbers, because the ego CAV cannot communication with other CAVs under the fully-distributed control plane.
To this end, we predict the number of point clouds by leveraging the LiDAR scanning mechanism, which generally perceives 360-degree surroundings with even distributed points.


Third, we iteratively remove multi-viewed point clouds for each sub-space of the object.
We build a directional star-like graph ($V$, $E$), where the center vertex is the current detected object.
Besides, vertices $V$ also include other potential multi-viewed CAVs, and edges $E$ denote the predicted number of point clouds.
In each iteration, we sum up the total number of point clouds of existing edges $E$, and compared with the given threshold of space density. 
The edge $e$ with the least edge value (i.e., the number of point cloud) will be removed, if the sum of remaining edges is still larger than the threshold.
The above process iterates until the threshold is met.

Note that, although the object selection is optimized in each CAV independently, all CAVs share the same global information (e.g., trajectory modeling and point cloud prediction), from the periodical updates of AdaMap.
As a result, the optimization result of object selection in all CAVs would be nearly the same, even if under the fully distributed scenarios in AdaMap.

\subsubsection{Approximated Gradient Descent}
Here, we focus on optimizing the RF for only selected objects in individual CAVs.
The challenges of RF optimization lie in 1) the percentile constraint of latency, and 2) the probabilistic fidelity and latency models.

To tackle the percentile constraint Eq.~\ref{eq:const_latency_individual} in the subproblem, we design a two-layer iterative approach based on the Lagrangian primal-dual method~\cite{boyd2004convex}.
The basic idea is to adaptively weight and incorporate the constraint into the objective function Eq.~\ref{eq:objective_function_individual}, which converts the constrained subproblem $\mathbb{P}_1$ into a series of unconstrained subproblems.
First, we relax the discrete variables $\mathcal{R}_i$ into continuous variables denoted by $\hat{\mathcal{R}}_i$.
Second, we build the Lagrangian as 
\begin{equation}
    \mathcal{G}(\hat{\mathcal{R}}_i, \lambda) = \mathbb{E}[F_i] + \lambda \left( Prob(L_i \le H) - p \right),
    \label{eq:lagrangian}
\end{equation}
where a non-negative multiplier $\lambda$ is introduced to weight the constraint.
Then, the subproblem $\mathbb{P}_1$ can be addressed by alternatively addressing the primal and dual subproblem.
Specifically, the dual subproblem is expressed as
\begin{equation}
    \lambda^* = \arg\max \limits_{ \lambda \ge 0} \mathcal{G}(\hat{\mathcal{R}}_i, \lambda).
\end{equation}
In contrast, the primal subproblem is expressed as
\begin{align}
\label{eq:primal_problem}
\hat{\mathcal{R}}_i^* & = \arg\max\limits_{\hat{\mathcal{R}}_i \in (\ref{eq:const_action_r})}  \mathcal{G}(\hat{\mathcal{R}}_i, \lambda),
\end{align}
where $\hat{\mathcal{R}}_i^*$ denotes the optimal continuous RFs. 
In the outer layer, the dual subproblem is solved under the given RFs, by using the sub-gradient descent~\cite{boyd2004convex}.
In the inner layer, the primal subproblem is solved under the given the multiplier $\lambda$.
As the convergence of the two-layer iteration, the obtained continuous RFs will be discretized into the discrete values in $[r_{\min}, .., r_{\max}]$.

Due to the probabilistic parts in the above primal subproblem, its falls into the field of stochastic programming~\cite{shapiro2007tutorial}.
We observe that these probabilistic parts are dependent on only the representation module, which is generally offline trained before its real-world deployment in AdaMap.
Hence, we can offline evaluate the representation module extensively to peek at the uncertainty of the primal subproblem.
Specifically, we vary the RFs, measure the resulted probabilistic fidelity and latency, and collect the measurement datasets, as shown in Fig.~\ref{fig:RF_context}.
Given the collected datasets, we solve the primal subproblem by using gradient descent methods, which is described as follows.
First, we select arbitrary initial RFs.
Second, we deviate the current RFs and sample the corresponding probabilistic parts (i.e., fidelity $F_i$ and latency $L_i$) from the measurement datasets.
Third, we build an approximated local model for both fidelity and latency functions (i.e., $F_i$ and $L_i$) with respect to RFs (i.e., $\mathcal{R}_i$) by using compute-efficient linear regression techniques.
Fourth, we calculate the gradients based on the approximated models and use gradient descent to conduct a one-step update to the current RFs.
The above process will be iterated toward its convergence.
The pseudo code of the approximated gradient descent algorithm is shown in Alg.~\ref{alg:1}.

\begin{algorithm}[!t]
	\caption{Approximated Gradient Descent}\label{alg:1}
    \KwIn{$\hat{\mathcal{A}}_i$, $r_{\min}$, $r_{\max}$ }
    \KwOut{$\hat{\mathcal{R}}_i$}

    Offline collect measurement datasets $\mathcal{D}$\;
    Initialize multiplier $\lambda$ and start point $\hat{\mathcal{R}}_i$\;
    
    \For{$m=1,2,...$}{
        $/**$ \textit{solve primal subproblem} $**/$\;
        \For{$n=1,2,...$}{
            Deviate $\hat{\mathcal{R}}_i$ via adding random noises $N_o$\;
            Sample fidelity and latency from datasets $\mathcal{D}$, under $\hat{\mathcal{A}}_i$ and $\hat{\mathcal{R}}_i + N_o$\; 
            Build linear regression models for fidelity $F_i$ and latency $L_i$\; 
            Calculate gradients based on regressed models\;
            Update $\hat{\mathcal{R}}_i$ via one-step gradient descent\;
        }
        $/**$ \textit{solve dual subproblem} $**/$\;
        Calculate the achieved percentile latency\;
        Update multiplier $\lambda$ via sub-gradient descent\;
    }
    Discretize $\hat{\mathcal{R}}_i$ to $\mathcal{R}_i$ in $[r_{\min}, r_{\max}]$\;
	\Return{$\mathcal{R}_i$}\; 
\end{algorithm}



	  
   
        

    
          


        	 
	

\section{System Implementation}

\begin{figure*}[!t] 
\captionsetup{justification=centering}
  \begin{minipage}[t]{0.32\textwidth}
    \centering
    \includegraphics[width=\textwidth]{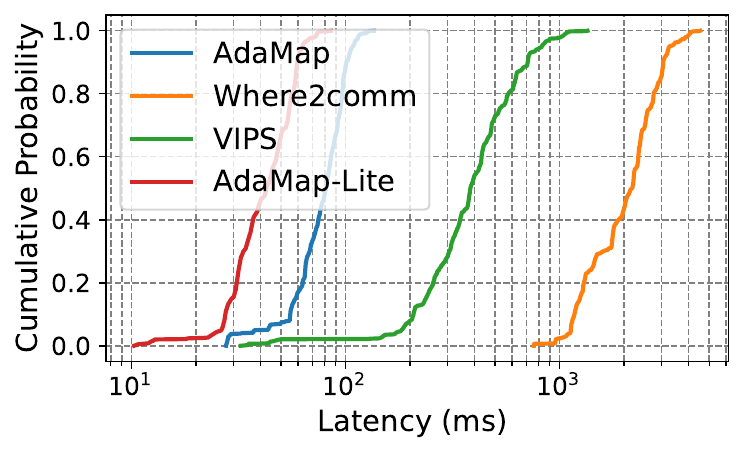}
    \captionof{figure}{\small Cumulative probability of latency.}
    \label{fig:LatencyCMP}
  \end{minipage}
  \begin{minipage}[t]{0.32\textwidth}
    \centering
    \includegraphics[width=\textwidth]{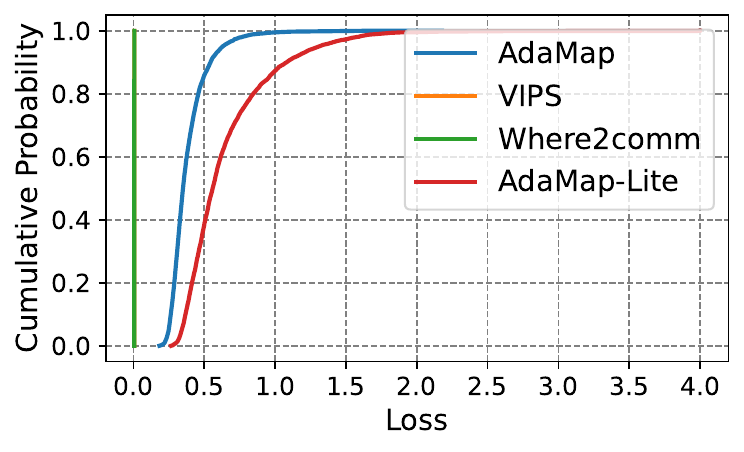}
    \captionof{figure}{\small Cumulative probability of losses.}
    \label{fig:FidelityCMP}
  \end{minipage}
  \begin{minipage}[t]{0.32\textwidth}
    \centering
    \includegraphics[width=\textwidth]{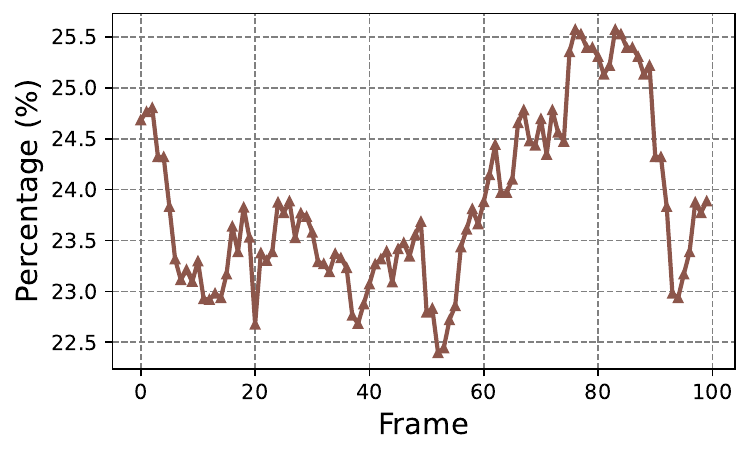}
    \captionof{figure}{\small Detailed object selection under AdaMap.}
    \label{fig:detail_selection}
  \end{minipage}
\end{figure*}




We implement AdaMap on a desktop computer with Intel i7 CPU and RTX 3090 GPU under Ubuntu 20.04.
The AdaMap system is implemented with mainly three parts, i.e., vehicles, the simulated network, and servers, where only object descriptors will be transmitted.
Due to the intractable complexity of conducting real-world experiments for hundreds of realistic vehicles, we choose to run all the data plane on the same desktop computer.
Given the limited computation capacity of the desktop computer (e.g., CPU cores and GPU memory), we execute the data plane of all CAVs sequentially and collect their vehicle-part computation latencies and transmission data sizes individually.
At runtime, the simulated network takes the necessary information (e.g., vehicle locations and transmission data sizes) as the input and simulates the transmission latency of individual CAVs.
Specifically, the simulated network is achieved by using a 5G system-level simulator~\cite{oughton2019open}, where the urban micro (UMi, Street Canyon) radio channel is adopted and the base station (BS) is located in the center of all CAVs by default\footnote{The evaluation results of different vehicle-BS distance can be found in Fig.~\ref{fig:fidelity_distance} and Fig.~\ref{fig:latency_distance} in Sec.~\ref{sec:evaluation}.}.
Then, the servers continue the post-processing of the data plane accordingly.
Finally, the latency of vehicle-part computation, network transmission, and server-part computation are aggregated to calculate the end-to-end latency of individual CAVs.
To implement AdaMap, we have finished more 4000 lines of codes in Python.
Note that, the AdaMap implementation is ready for real-world deployments, where only trivial networking configurations (e.g., pairing IP addresses) are needed to deploy on realistic cars and servers.

In the data plane, we use state-of-the-art PointPillars~\cite{lang2019pointpillars} as the 3D object detector. 
The threshold in our hybrid localization module is empirically tested and set to be 0.5m.
To achieve the representation module, we adopt the PointNet~\cite{qi2017pointnet} as the base encoder, and multiple fully-connected layers for the base decoder.
The early-exit tunnels are implemented with 256x128x64x32 and 32x64x128x256, where all input point clouds are resampled to 1024 points.
In other words, four discrete early-exit tunnels are available and thus the action space of RFs are [4, 8, 16, 32, 64]. 
In the loss function of the representation module, the weight $\beta=0.0001$. 

In the control plane, the percentile $p=0.99$ and the latency threshold $H=100ms$ for meeting the requirement of 10Hz LiDARs. 
We use 4 partitions for individual CAVs, where the threshold for the space density is 1024 per sub-space.
This is based on our experimental measurement in CARLA, where 4096 points are generally sufficient to represent vehicles.

To collect the dataset for AdaMap evaluation, we use CARLA simulator under the $town5$ with 150 vehicles and automatic navigation.
We collect more than 100 time frames at 10Hz interval.
The dataset includes the raw data of LiDAR and the ground-truth transformation matrix $T_{l2g}$ for each vehicle at all time frames, at the total size of 82GB.
The specification of LiDARs are 64-thread, 2.4M/s data rate, and 50m sensing range, where LiDARs are mounted on the 1 meter above the vehicle's roof.

We compare \textit{AdaMap} with the following state-of-the-art cooperative perception systems: 
\begin{itemize}
    \item VIPS~\cite{shi2022vips}: VIPS is a state-of-the-art vehicle-to-infrastructure (V2I) solution for cooperative perception. Its basic idea is to transmit only the interested objects to the ego vehicle, without point cloud compression. For a fair comparison, we implement VIPS to use off-the-shelf compression library (i.e., Zstandard) to transmit only detected objects (rather than the whole raw point cloud) of individual vehicles.
    \item Where2comm~\cite{hu2022where2comm}: Where2comm is a representative vehicle-to-vehicle (V2V) solution for extending the perception range of vehicles. Its basic idea is to share only interested areas (e.g., vision occlusion) between vehicles.
    For a fair comparison, we implement Where2comm to use Zstandard compression library to transmit point clouds of the whole blind spot among individual vehicles.
    \item AdaMap-Lite: The AdaMap-Lite adopts the data and control plane of AdaMap, but it selects all the detected objects and uses the maximum RF at all time frames.
\end{itemize}

\section{Performance Evaluation}
\label{sec:evaluation}

In this section, we evaluate AdaMap to answer the following questions. 1) How AdaMap performs compared to state-of-the-art solutions? 2) How AdaMap scales under different roadway traffic and network conditions? 3) How AdaMap adapts to network dynamics? 4) How individual components in the data and control plane impact the overall performance of AdaMap? 
In particular, we evaluate AdaMap with two key metrics, including percentile end-to-end latency and average reconstruction loss.
Note that, as existing solutions adopt the lossless point cloud compression library, their reconstruction losses are always zeros.

\begin{figure*}[!t] 
\captionsetup{justification=centering}
  \begin{minipage}[t]{0.32\textwidth}
    \centering
    \includegraphics[width=\textwidth]{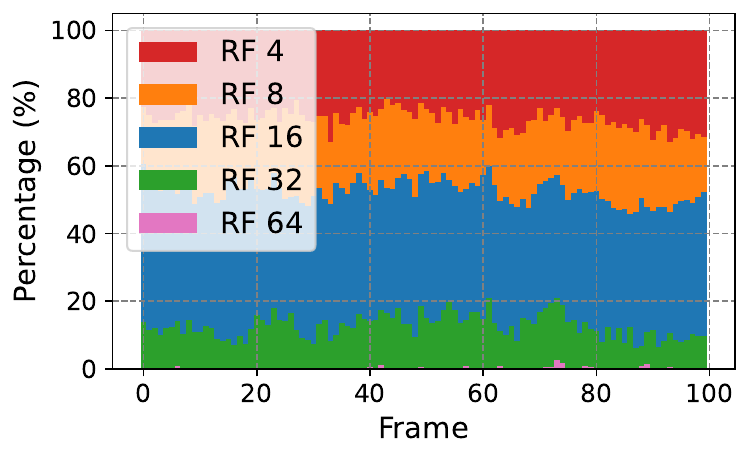}
    \captionof{figure}{\small Detailed RF optimization under AdaMap.}
    \label{fig:deatil_RF}
  \end{minipage}
  \begin{minipage}[t]{0.32\textwidth}
    \centering
    \includegraphics[width=\textwidth]{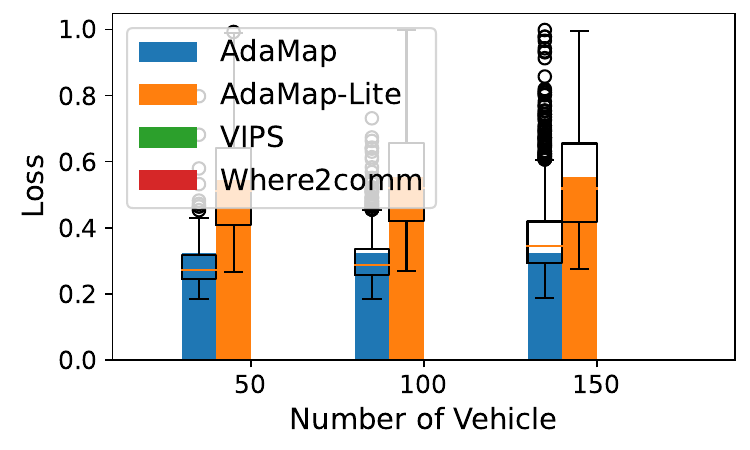}
    \captionof{figure}{\small Loss performance under different number of CAVs.}
    \label{fig:fidelity_car_num}
  \end{minipage}
  \begin{minipage}[t]{0.32\textwidth}
    \centering
    \includegraphics[width=\textwidth]{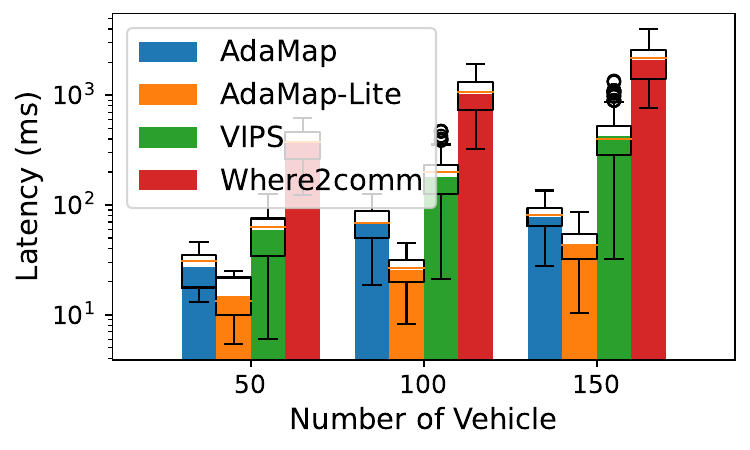}
    \captionof{figure}{\small Latency performance under different number of CAVs.}
    \label{fig:latency_car_num}
  \end{minipage}
\end{figure*}

\begin{figure*}[!t] 
\captionsetup{justification=centering}
  \begin{minipage}[t]{0.32\textwidth}
    \centering
    \includegraphics[width=\textwidth]{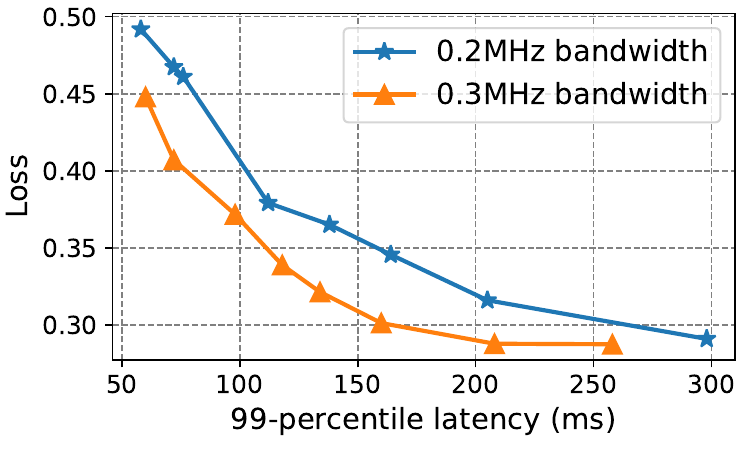}
    \captionof{figure}{\small Pareto boundary of AdaMap.}
    \label{fig:pareto}
  \end{minipage}
  \begin{minipage}[t]{0.32\textwidth}
    \centering
    \includegraphics[width=\textwidth]{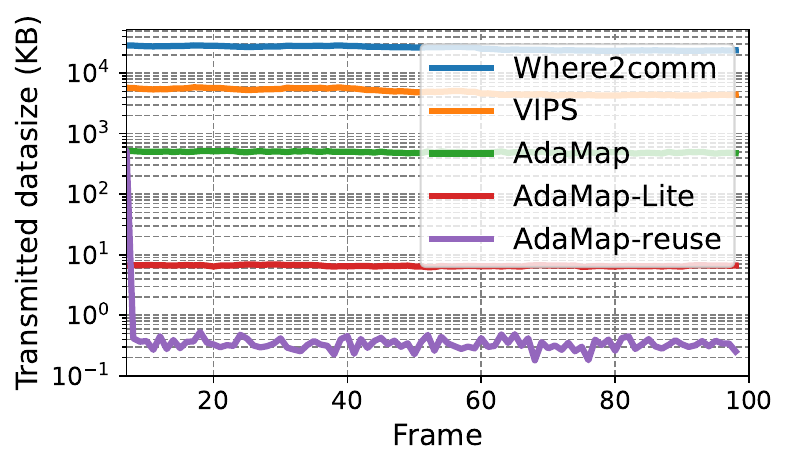}
    \captionof{figure}{\small Transmission size under different solutions.}
    \label{fig:over_time}
  \end{minipage}
  \begin{minipage}[t]{0.32\textwidth}
    \centering
    \includegraphics[width=\textwidth]{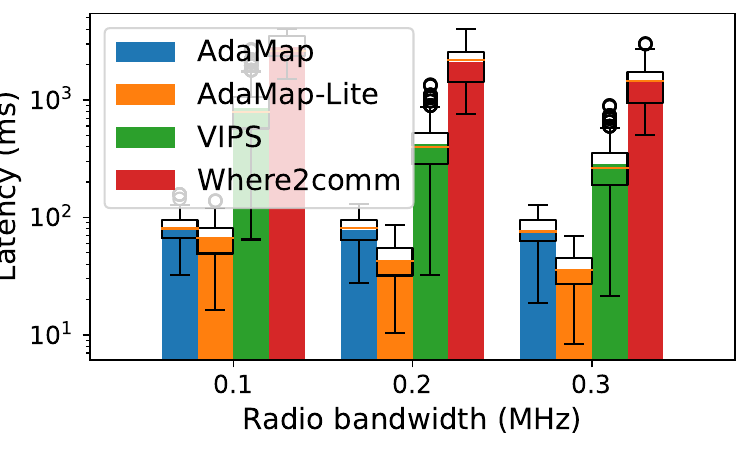}
    \captionof{figure}{\small Latency performance under different radio bandwidth.}
    \label{fig:latency_bandwidth}
  \end{minipage}
\end{figure*}

\subsection{Overall Performance}

\begin{figure*}[!t] 
\captionsetup{justification=centering}
  \begin{minipage}[t]{0.32\textwidth}
    \centering
    \includegraphics[width=\textwidth]{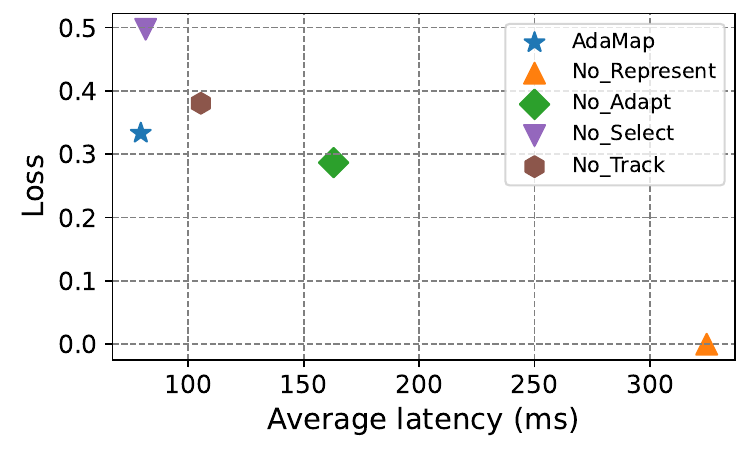}
    \captionof{figure}{\small Impact of individual components in AdaMap.}
    \label{fig:individual_components}
  \end{minipage}
  \begin{minipage}[t]{0.32\textwidth}
    \centering
    \includegraphics[width=\textwidth]{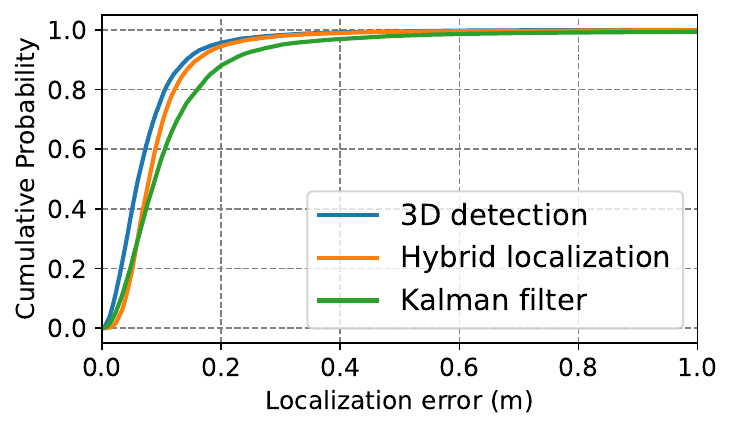}
    \captionof{figure}{\small Localization errors under different methods.}
    \label{fig:localization_error}
  \end{minipage}
  \begin{minipage}[t]{0.3\textwidth}
    \centering
    \includegraphics[width=\textwidth]{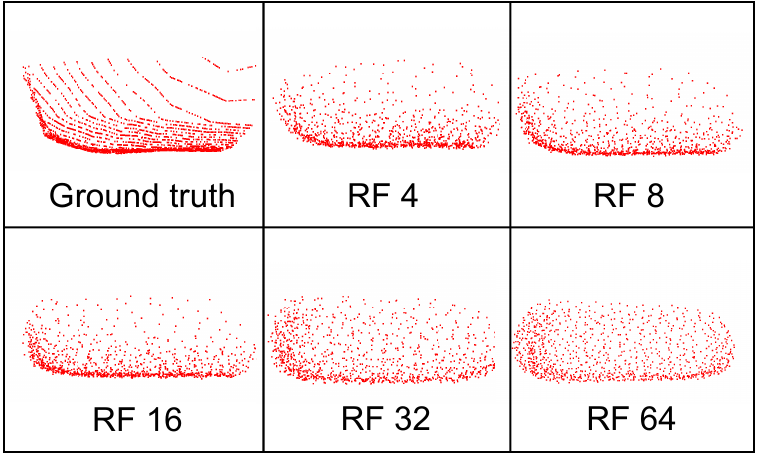}
    \captionof{figure}{\small Example of reconstructed point clouds.}
    \label{fig:visual_effect}
  \end{minipage}
\end{figure*}

We evaluate the overall latency and loss performance, which indicates the efficacy of cooperative perception systems.
Fig.~\ref{fig:LatencyCMP} and Fig.~\ref{fig:FidelityCMP} show the cumulative probability of end-to-end latency and reconstruction loss achieved by different solutions under 150 CAVs.
For the latency performance, AdaMap reduces 9.8x and 49x on average as compared to VIPS and Where2comm, respectively.
Under limited networking resources (i.e., 200KHz radio bandwidth), existing solutions consume 0.42 $\pm$ 1.5s and 2.12 $\pm$ 2.4s to complete the perception sharing for all CAVs, which cannot be tolerated in real world deployment.
In contrast, AdaMap achieves 70ms latency on average, where 
89.4\% CAVs have less than 100ms end-to-end latency. 
Note that, we evaluate AdaMap with $p=0.99$ by default, which indicates that the percentile latency constraint in Eq.~\ref{eq:const_latency} is not fully satisfied here.
We found the key reason is the limited number of discrete RFs in AdaMap, where the data plane supports only a total of 5 RFs.
When the continuous RFs are discretized, the achieved optimality is compromised to a certain extent.
In other words, this gap would be bridged by developing more fine-grained RFs in real-world deployment.
For the loss performance, AdaMap achieves 0.37 reconstruction loss on average, where 95\% CAVs are with less than 0.63 loss.
Note that, Table~\ref{tab:loss_under_rf} shows the reconstruction loss under RF 64 is 0.48 on average, where the resulted visual effects are generally acceptable for autonomous driving pipeline as illustrated in Fig.~\ref{fig:visual_effect}.
These results verify that AdaMap can achieve high-scalable real-time cooperative perception by using very limited networking resources.


Besides, we dissect AdaMap in details regarding how the control plane optimizes the object selections and representation factors (RFs).
Fig.~\ref{fig:detail_selection} and Fig.~\ref{fig:deatil_RF} show the percent of selected objects and optimized RFs over 100 time frames.
It can be seen that, AdaMap adaptively selects 23.6\% objects on average, which indicates that a large portion of multi-viewed point clouds of objects are duplicated in terms of contributing to the overall fidelity.
Hence, we observe that the transmission size of all CAVs are significantly reduced, which generally improves the latency performance under fixed networking resources.
In addition, we observe that optimized RFs in AdaMap are distributed from $r_{\min}=4$ to $r_{\max}=64$ in Fig.~\ref{fig:deatil_RF}.
Generally, AdaMap would choose the RF with the highest compression ratio for trading off the latency performance under heavy traffic scenarios, such as traffic jam. 
However, AdaMap rarely uses RF 64 (only 0.55\%) but uses RF 32 at the probability of 20.9\%.
This might be attributed to 1) the mechanism of evenly vehicle dispatch in CARLA simulator generally leads to non-heavy traffic scenarios; 2) the object selections in AdaMap reduce the number of objects to be transmitted before the RF optimization.



\subsection{Scalability}

We evaluate the scalability of AdaMap by varying the number of CAVs and percentile latency requirement.
Fig.~\ref{fig:fidelity_car_num} and Fig.~\ref{fig:latency_car_num} show the reconstruction loss and end-to-end latency achieved by different solutions under different number of CAVs.
It can be seen that, AdaMap can always maintain the percentile latency, while existing solutions experience exponential increment on their latency performances.
In particular, AdaMap achieves 27.1ms latency on average when there are 50 CAVs, where the optimized RFs are mostly focused on RF 4 (98.6\%).
Moreover, AdaMap reduces the average reconstruction loss by 23.8\% as compared to that in high traffic (150 CAVs).

\begin{figure*}[!t] 
\captionsetup{justification=centering}
\begin{minipage}[t]{0.32\textwidth}
    \centering
    \includegraphics[width=\textwidth]{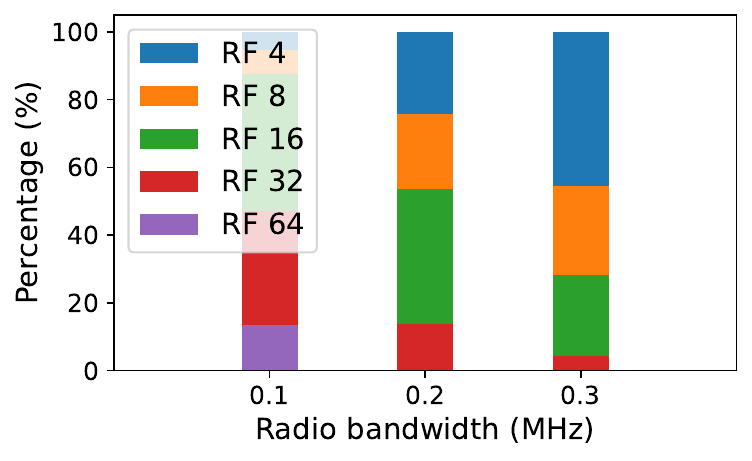}
    \captionof{figure}{\small Detailed RF optimization under different radio bandwidth.}
    \label{fig:intelligent_RF_bandwidth}
  \end{minipage}
  \begin{minipage}[t]{0.32\textwidth}
    \centering
    \includegraphics[width=\textwidth]{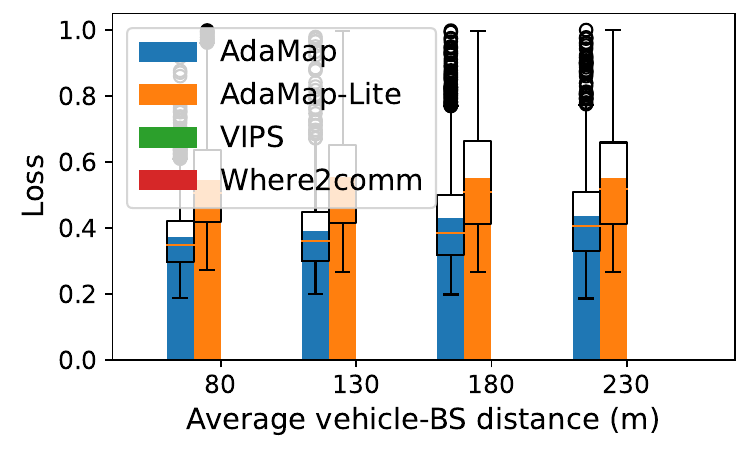}
    \captionof{figure}{\small Loss performance under different average vehicle-BS distances.}
    \label{fig:fidelity_distance}
  \end{minipage}
  \begin{minipage}[t]{0.32\textwidth}
    \centering
    \includegraphics[width=\textwidth]{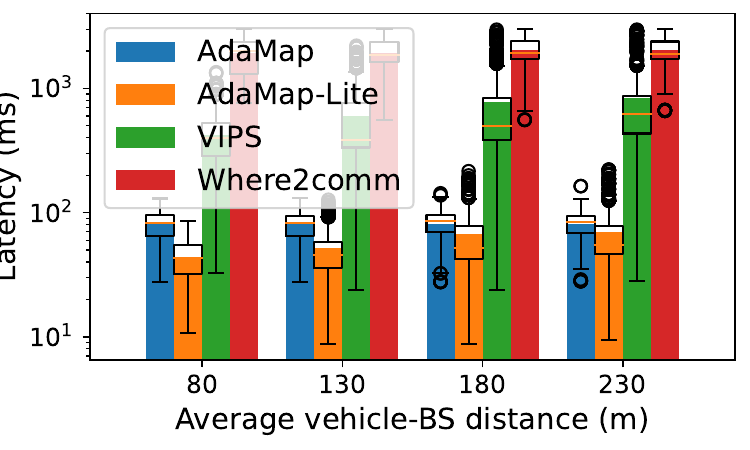}
    \captionof{figure}{\small Latency performance under different average vehicle-BS distances.}
    \label{fig:latency_distance}
  \end{minipage}
\end{figure*}

In Fig.~\ref{fig:pareto}, we obtain the Pareto boundary of AdaMap by varying the 99-percentile latency requirement under 150 CAVs.
We observe that AdaMap can achieve the lowest 99-percentile latency of 57.7ms at the cost of 0.49 reconstruction loss, and the lowest loss of 0.29 when the latency requirement is loosen to 298ms. 
To push the boundary towards even lower loss and latency, one of feasible approaches is to increase available networking resources, such as radio bandwidth.
In particular, the achieved 99-percentile latency of 138ms under 200KHz can be further reduced to 102ms with the same average loss of 0.36, when the radio bandwidth is expanded to 300KHz.
Fig.~\ref{fig:over_time} shows the total transmission size of all CAVs under different solutions in 100 frames.
The transmission size is determined by not only the number of total selected objects but also the average size of each objects.
In AdaMap, we transmit only selected objects, where each object is compressed by using the representation module with high compression ratios. 
Hence, AdaMap reduces on average 89.7\% and 98.12\% transmission size as compared to VIPS and Where2comm, respectively.
In particular, benefit from the object matching in AdaMap, we can track all the objects in the global map.
Hence, considering the reusing of historical point cloud of objects, we observe that AdaMap-Reuse further reduces 98.1\% as compared to original AdaMap.
Besides, AdaMap-Reuse obtains 97.8\% and 99.95\% reduction on the transmission size as compared to VIPS and Where2comm, respectively.
These results demonstrate that AdaMap is highly scalable to tackle large-scale traffic scenarios.

\subsection{Adaptability}


We evaluate the adaptability of AdaMap by varying radio bandwidth and channel qualities.
Fig.~\ref{fig:latency_bandwidth} shows the impact of radio bandwidth on the latency of different solutions.
With 100KHz more radio bandwidth, VIPS and Where2comm reduce 25.3\% and 50.1\% average latency than that under 200KHz radio bandwidth, respectively.
In contrast, AdaMap and AdaMap-Lite maintain the percentile latency by adaptively optimizing RFs, where the reconstruction loss is reduced by 23.2\% on average under 100KHz more radio bandwidth.
Fig.~\ref{fig:intelligent_RF_bandwidth} shows the detailed RF optimization under different radio bandwidth. 
We see that, the average RF is increased from 13.5 when the bandwidth is 200KHz to 9.1 under 300KHz bandwidth.
Under more radio bandwidth, AdaMap can intelligently adapt to available network resources to reduce the reconstruction loss and improve the overall fidelity.

Fig.~\ref{fig:fidelity_distance} and Fig.~\ref{fig:latency_distance} depict the impact of radio channel qualities on the performance of different solutions.
Specifically, this result is collected by changing the location of the base station (BS) in the simulated network to vary the distance between vehicles and base stations.
The larger vehicle-BS distance generally indicates lower achievable wireless data rates and higher fluctuations in the meantime.
We observe that, AdaMap can always satisfy the percentile latency requirements, at the cost of increasing the average reconstruction loss.
These results verify that AdaMap can effectively adapt to time-varying network dynamics.

\subsection{Design of Data Plane}
We evaluate the data plane in terms of localization error and reconstruction loss.
In Fig.~\ref{fig:localization_error}, we show the cumulative probability of localization error, which is measured by the Euclidean distance between the predicted and ground-truth location.
It can be seen that, our hybrid localization module achieves 0.19m localization error with the probability of 95\%.
As compared to purely object detection, our hybrid localization module obtains 96.1\% reduction on the inference time at the cost of only 0.02m localization loss.

Besides, we illustrate the visual effects of the point cloud reconstructed by the representation module (see Sec.~\ref{sec:data_plane}) under different representation factors (RFs) in Fig.~\ref{fig:visual_effect}.
It can be seen that, the reconstructed point cloud under RF 4 is very close to the original point cloud structurally.
Although the reconstructed point cloud under RF 64 is relatively far from the original, we believe it still provides sufficient structural details for following autonomous driving pipeline, such as 3D scene understanding (more discussions in Sec.~\ref{sec:discussion}).
Note that, we implemented the representation module with relatively small DNNs, which indicates that its reconstruction loss could be further reduced by using larger DNNs with efficient and sophisticated neural network architectures~\cite{liu2020morphing}.

In Table~\ref{tab:module_time}, we show the computing time of individual data plane modules in AdaMap.
With the unique design in the hybrid localization module, AdaMap only spends negligible delay on localizing interested objects.
In particular, the most time-consuming module is the representation module (1.72ms on average), which involves the computation of DNNs. 
Note that, all these modules are very compute-efficient and can be completed in a few milliseconds, which justifies the computation efficiency of data plane in AdaMap.


\begin{table}[!t]
    \captionof{table}{Computation time of data plane modules}
    \begin{tabular}[b]{c|c c}\hline
       \textbf{Modules}               &  \textbf{Inference time (ms)}      \\ \hline
       \textbf{ Object Localization}              & 0.061 $\pm$ 0.023 \\ 
       \textbf{ Representation}                   & 1.72 $\pm$ 0.53\\
       \textbf{ Perspective Transformation}                   & 0.006 $\pm$ 0.012 \\
       \textbf{ Object Matching}            & 0.014 $\pm$ 0.023  \\ \hline
    \end{tabular}
\label{tab:module_time}
\end{table}

\subsection{Impact of Individual Components}


We evaluate individual components in contributing to the overall performance of AdaMap.
Overall, we observe that AdaMap achieves the best performance in terms of meeting the percentile latency while reducing the reconstruction loss.
Without using the representation module in the data plane,  \textit{No\_Represent} relies on existing off-the-shelf compression, which increases 4.63x average latency with no reconstruction loss.
Without using the object selection in the control plane, \textit{No\_Select} transmits all detected objects, which results in higher optimized RFs and higher reconstruction loss.
Without using the hybrid detection and tracking in the data plane, \textit{No\_Track} relies on only object detection to localize interested objects, which squeezes the latency allowance for following data plane modules, and leads to worse performance on both latency and loss.
Besides, \textit{No\_Adapt} uses only RF 4 all the time, which cannot adapt to network dynamics and thus fails the percentile latency requirement. 



\vspace{0.05in}
\section{Related Work}
\vspace{0.05in}
\textbf{Cooperative perception} has been increasingly studied to augment the perception for the ego vehicle towards safe autonomous driving.
AVR~\cite{qiu2018avr} shares raw point clouds among vehicles to extend their sensing ranges and reduces the transmission size by selecting only dynamic objects and motion vectors.
F-cooper~\cite{chen2019f} introduced the first feature-level fusion solution to combine both voxels and features of different LiDAR point clouds, which improves the accuracy of 3D object detection in the ego vehicle.
Recently, infrastructure-assisted solutions emerge to tackle the exponential increment of transmission data size under vehicle-to-vehicle (V2V) approaches.
EdgeSharing~\cite{liu2021edgesharing} leveraged edge computing servers to maintain a global map, that aggregates various crowdsourced data (e.g., raw point clouds and features), to assist the localization, detection, and matching for individual CAVs.
To improve the scalability of cooperative perception, multiple works (e.g., AutoCast~\cite{qiu2022autocast}, VIPS~\cite{shi2022vips}, and Where2comm~\cite{hu2022where2comm}) further exploited the spatial correlation and individual needs among CAVs to reduce the transmission data size.
However, these works cannot scale in large-scale deployment with hundreds of CAVs (if not more), such as block and community scales, in terms of networking resource demand and end-to-end latency assurance.

\textbf{Edge computing} is expected to shape and revolutionize the future automotive industry with low-latency accessible resources, where computation offloading provides promising computation acceleration~\cite{zhang2018joint, chen2021drl, 10139782, 9051801}.
DeepDecision~\cite{ran2018deepdecision} outlined an optimization framework to determine the offloading strategy for compute-intensive workloads, while balancing the detection accuracy, energy consumption, and network data usage.
Edge-SLAM~\cite{ben2022edge} is designed to split the computation modules of visual simultaneous localization and mapping(Visual-SLAM) between mobile devices and edge servers, to accelerate the overall computation by leveraging the high computation capacity at edge servers.
LiveMap~\cite{liu2021livemap} introduced a new collaborative computing framework to intelligently and adaptively place partial computation workloads in vehicles and edge servers, during the construction of real-time dynamic map at the edge. 
However, existing works are mostly focused on centralized optimization with deterministic system models, which incurs significant communication delay and overhead under large-scale cooperative perception scenarios. 

\vspace{0.05in}
\section{Discussion}
\label{sec:discussion}
\vspace{0.05in}
The main focus of this work lies in the fundamental trade-offs between the latency and fidelity in large-scale cooperative perception systems.
One of the key designs is the representation module for lossy compression of point clouds, which allows adaptive balancing between latent dimensions and reconstruction losses at runtime. 
From the visualization results in Fig.~\ref{fig:visual_effect}, we believe that our implementation of the representation module remains noticeable rooms for further improvement, in terms of reducing reconstruction losses.
We envision two potential directions of improving its implementation as follows.
First, the design of its loss function might be further enhanced to balance local detail and global similarity, e.g., linear searching of the weight factor $\beta$.
Second, the design of its neural network architecture could be further optimized to improve the training performance, e.g., neural architecture search (NAS) and automatic hyper-parameter tuning. 

In addition, one of the limitations of the representation module is the unpredictable (e.g., potentially catastrophic) reconstruction losses under unseen point clouds in the training dataset.
This limitation may be alleviated by 1) building a complete training dataset as much as possible to cover diversified point clouds; 2) enhancing its generalization by adopting robust and adversarial training techniques; 3) predicting the reconstruction loss (refer to Fig.~\ref{fig:RF_context}) and resort to lossless compression techniques if the predicted loss is too high.

Furthermore, the evaluation of this work is based on accurate local-to-global conversion matrix and non-contaminated sensory data, which might not be available in real-world deployment scenarios. 
The robustness of cooperative perception systems would need further extensive evaluation under real-world experiments, such as the error propagation under detection errors and localization shifts.
Besides, the corresponding countermeasures demand for further exploratory study, experimental testing, and real-world validation.




\vspace{0.05in}
\section{Conclusion}
\vspace{0.05in}
In this paper, we presented AdaMap, a new high-scalable real-time cooperative perception system.
We designed a tightly coupled data plane and control plane to achieve AdaMap.
In the data plane, we designed the hybrid localization module to accelerate the object localization, and the flexible representation module to trade acceptable reconstruction losses for higher compression ratios.
In the control plane, we designed the object selection method to mitigate the excessive multi-viewed point clouds, and the approximated gradient descent algorithm to online optimize the representation factor (RF) under time-varying network dynamics.
We evaluated AdaMap in an emulation platform to show its scalability, adaptability and computation efficiency.
The evaluation results showed that, AdaMap can assure percentile latency requirements and reduce up to 49x transmission data size at the cost of 0.37 reconstruction loss than state-of-the-art solutions.
We believe that, AdaMap sheds the light on large-scale real-time cooperative perception deployment in future scenarios.

\vspace{0.05in}
\section*{Acknowledgement}
\vspace{0.05in}
This work is partially supported by the US National Science Foundation under Grant No. 2321699.


\bibliographystyle{acm}
\bibliography{ref/reference.bib, ref/qiang.bib}

\end{document}